\documentclass{article}

\usepackage{ijcai24}

\usepackage{times}
\usepackage{soul}
\usepackage{url}
\usepackage[hidelinks]{hyperref}
\usepackage[utf8]{inputenc}
\usepackage[small]{caption}
\usepackage{graphicx}
\usepackage{amsmath}
\usepackage{amsthm}
\usepackage{booktabs}
\usepackage{algorithm}
\usepackage{algorithmic}
\usepackage[switch]{lineno}

\usepackage{algorithm}
\usepackage{algorithmic}
\usepackage{multirow}
\usepackage{booktabs}
\usepackage{url}
\usepackage{amsmath}
\usepackage{amsfonts}
\usepackage{threeparttable}
\usepackage{nicefrac}
\usepackage{adjustbox}
\usepackage{subcaption}

\usepackage{color}

\title{Spatial-Temporal-Decoupled Masked Pre-training for Spatiotemporal Forecasting}

\author{
Haotian Gao$^{1,2}$\and
Renhe Jiang$^{1}$\thanks{Corresponding author.}\and
Zheng Dong$^{2}$\and
Jinliang Deng$^{3}$ \and
Yuxin Ma$^{2}$ \and
Xuan Song$^{2}$
\affiliations
$^1$The University of Tokyo,
$^2$Southern University of Science and Technology \\
$^3$University of Technology Sydney
\emails
    gaoht6@outlook.com, 
    jiangrh@csis.u-tokyo.ac.jp, 
    zhengdong00@outlook.com\\
    jinliang.deng@student.uts.edu.au, 
    mayx@sustech.edu.cn, 
    songx@sustech.edu.cn
}

\begin{document}

\maketitle
\begin{abstract}
Spatiotemporal forecasting techniques are significant for various domains such as transportation, energy, and weather. Accurate prediction of spatiotemporal series remains challenging due to the complex spatiotemporal heterogeneity. In particular, current end-to-end models are limited by input length and thus often fall into \textbf{\textit{spatiotemporal mirage}}, i.e., similar input time series followed by dissimilar future values and vice versa. To address these problems, we propose a novel self-supervised pre-training framework Spatial-Temporal-Decoupled Masked Pre-training (\textbf{STD-MAE}) that employs two decoupled masked autoencoders to reconstruct spatiotemporal series along the spatial and temporal dimensions. Rich-context representations learned through such reconstruction could be seamlessly integrated by downstream predictors with arbitrary architectures to augment their performances. A series of quantitative and qualitative evaluations on six widely used benchmarks (PEMS03, PEMS04, PEMS07, PEMS08, METR-LA, and PEMS-BAY) are conducted to validate the state-of-the-art performance of STD-MAE. Codes are available at \url{https://github.com/Jimmy-7664/STD-MAE}.
\end{abstract}

\section{Introduction}
Spatiotemporal data collected by sensor networks has become a vital area of research with many real-world applications. It benefits from extra spatial context like sensor locations and road networks that reveal dependencies between sensors. Consequently, a key distinction from typical multivariate time series is that spatiotemporal data exhibits spatiotemporal heterogeneity. Specifically, while time series vary across different locations (urban centers against suburban areas) and day types (weekday versus weekend), they demonstrate consistent, predictable patterns within similar contexts. Hence, accurately predicting spatiotemporal data hinges on effectively capturing such heterogeneity.
\begin{figure}[h]
  \centering
  \begin{subfigure}{1\linewidth}
    \includegraphics[width=\linewidth]{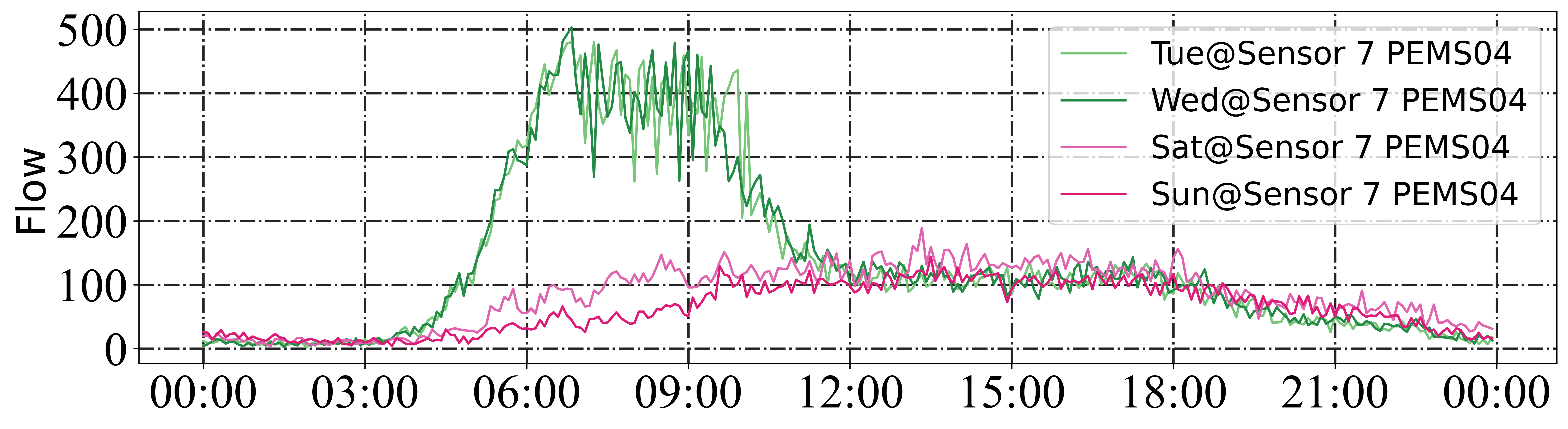}
    \caption{Temporal Heterogeneity}
    \label{fig1-sub1}
  \end{subfigure}
  \begin{subfigure}{1\linewidth}
    \includegraphics[width=\linewidth]{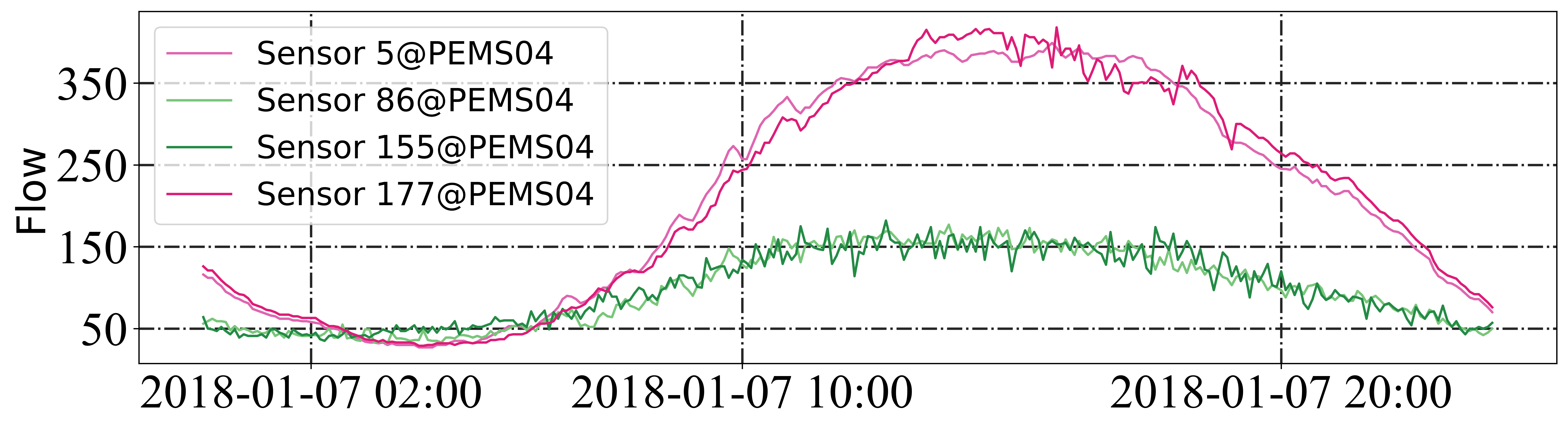}
    \caption{Spatial Heterogeneity}
    \label{fig1-sub2}
  \end{subfigure}
\begin{subfigure}{1\linewidth}
    \includegraphics[width=\linewidth]{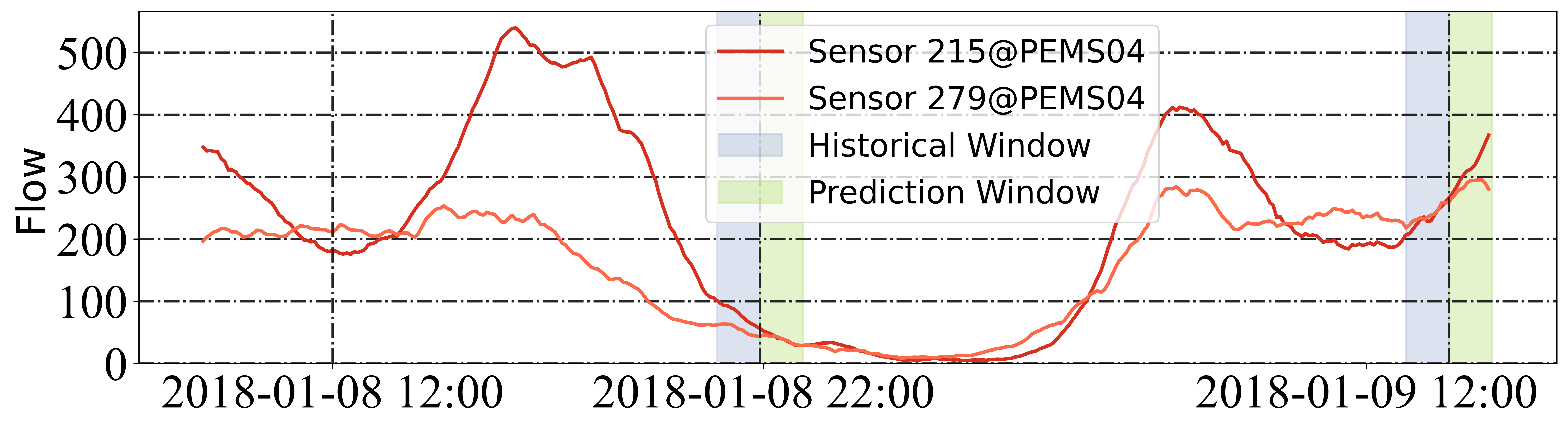}
    \caption{Spatiotemporal Mirage}
    \label{fig1-sub3}
  \end{subfigure}
  \caption{Illustration of Spatiotemporal Heterogeneity and Mirage}
  \label{fig1}
\end{figure}
Figure~\ref{fig1-sub1} shows the traffic flow of sensor 7 in a spatiotemporal dataset PEMS04, revealing distinct weekday and weekend patterns especially during peak hours. Weekdays experience a morning rush hour peak, whereas weekend is more evenly distributed without significant peaks, illustrating temporal heterogeneity in weekly patterns. Additionally, Figure~\ref{fig1-sub2} displays the traffic flow of sensor 5, 86, 155, and 177 during the same period. Sensor 5 and 177 exhibit distinct peaks and troughs, while sensor 86 and 155 remain relatively stable throughout the day, demonstrating spatial heterogeneity. When the data scale is small, the heterogeneity is distinctly visible as shown above. However, when months of data from hundreds of sensors are given, the spatial and temporal heterogeneity becomes highly mixed. Previous researchers have done various attempts for spatiotemporal forecasting: embedding GCN into TCN~\cite{yu2018spatio,wu2019graph} or RNN~\cite{li2018diffusion,bai2020adaptive}, or applying transformer~\cite{jiang2023pdformer,liu2023spatio} along spatiotemporal axes. But these models often have difficulty in distinguishing the spatial and temporal heterogeneity in a clear way. So learning clear heterogeneity is still the primary challenge for spatiotemporal forecasting. 

Moreover, most of the existing models are trained in an end-to-end manner. Due to their high model complexity, their input horizons are often restricted to a short value (usually 12 steps)~\cite{yu2018spatio,wu2019graph,song2020spatial}. This limitation will make the models suffer from an issue denoted as \textbf{\textit{spatiotemporal mirage}}: \textbf{\textit{i)  dissimilar input time series followed by similar future values}}; \textbf{\textit{ii) similar input time series followed by dissimilar future values}}. We illustrate this phenomenon by taking traffic flow of sensor 215 and 279 as an example in Figure~\ref{fig1-sub3}. In the late evening, the two sensors show divergent historical data trends but similar future flow. Conversely, in the afternoon, their historical data trends align closely, yet their future data differ dramatically. Essentially, the reason behind this is that existing models can only capture the fragmented heterogeneity instead of the complete one. Therefore, how to make these models robust on such spatiotemporal mirage issue is the second challenge. 

In this study, we focus our approach on learning \textbf{\textit{clear}} and \textbf{\textit{complete}} spatiotemporal heterogeneity through pre-training. In particular, masked pre-training has shown tremendous effectiveness in natural language processing~\cite{kenton2019bert} and computer vision~\cite{bao2021beit}. The core idea is to mask parts of the input sequence during pre-training, requiring the model to reconstruct the missing contents. In this way, the model learns context-rich representations which can augment various downstream tasks. Motivated by these benefits, we propose a novel spatial-temporal-decoupled masked pre-training framework called STD-MAE. It offers an effective and efficient solution for learning clear and complete spatiotemporal heterogeneity through pre-training. Such learned heterogeneity can be flawlessly integrated into downstream baselines to see through spatiotemporal mirages. In summary, our key contributions are as follows: 
\begin{itemize}
    \item We devise a pre-training framework on spatiotemporal data that can largely enhance downstream spatiotemporal predictors of arbitrary architectures without modifying their original structures. 
    \item We propose a novel spatial-temporal-decoupled masking strategy to effectively learn spatial and temporal heterogeneity by capturing long-range context across spatial and temporal dimensions. 
    \item  We validate STD-MAE on six benchmarks (PEMS03, PEMS04, PEMS07, PEMS08, METR-LA, and PEMS-BAY) with typical backbones. Quantitative enhancement on baselines highlights the exceptional performance of STD-MAE. Qualitative analyses demonstrate its power to capture meaningful long-range spatiotemporal patterns. 
\end{itemize}
\section{Related Work}
\subsection{Spatiotemporal Forecasting}
Spatiotemporal forecasting~\cite{jiang2021dl} aims to predict future spatiotemporal series by analyzing historical data. Early work mainly relied on traditional time series models~\cite{pan2012utilizing,stock2001vector}. To capture the complex temporal dependencies, RNNs~\cite{hochreiter1997long,chung2014empirical} and CNNs~\cite{oord2016wavenet} have gained popularity for better modeling spatiotemporal data and achieving improved predictions.

Nevertheless, these models overlook crucial spatial correlations, limiting predictive performance on networked road systems. To further capture spatiotemporal features jointly, some studies integrate Graph Convolutional Networks (GCNs) with temporal models~\cite{yu2018spatio,li2018diffusion}. Following this line of research, several novel spatiotemporal models have been proposed in recent years~\cite{wu2019graph,xu2020spatial,bai2020adaptive,cao2020spectral,wu2020connecting,guo2021hierarchical,deng2022multi,han2021dynamic,jiang2023spatio,deng2024disentangling}. 
Attention mechanisms~\cite{vaswani2017attention} have also profoundly influenced spatiotemporal forecasting. A series of transformers~\cite{xu2020spatial,jiang2023pdformer,liu2023spatio} are proposed and exhibit superior performance, highlighting their effectiveness in capturing the spatiotemporal relations. However, these end-to-end models only focus on short-term input that limits them to capture complete spatiotemporal dependencies. 

\subsection{Masked Pre-training} 
Masked pre-training has emerged as a highly effective technique for self-supervised representation learning in both natural language processing (NLP) and computer vision (CV). The key idea is to train models to predict masked-out parts of the input based on visible context. In NLP, approaches like BERT~\cite{kenton2019bert} use masked language modeling to predict randomly masked tokens with bidirectional context. Subsequent models~\cite{liu2019roberta,lan2019albert} introduced more efficient masking techniques and demonstrated performance gains from longer pre-training. In CV, similar masking strategies have been adopted. Methods like BEiT~\cite{bao2021beit} and Masked AutoEncoder (MAE)~\cite{he2022masked} mask out random patches of input images and do reconstruction based on unmasked patches. In both domains, masked pre-training produces substantial improvements on various downstream tasks.

Recently, many researchers have attempted to employ pre-training techniques on time series data to obtain superior hidden representations~\cite{nie2022time,shao2022pre,li2023ti}. However, these methods are either channel-independent or neglect pre-training in the spatial dimension. Our proposed STD-MAE introduces a novel spatial-temporal-decoupled masking strategy during pre-training. By masking separately on spatial and temporal dimensions, the learned representations can effectively capture the intricate long-range heterogeneity in spatiotemporal data.

\section{Problem Definition}
Spatiotemporal forecasting is a specialized multivariate time series forecasting problem. Given multivariate time series $X_{t-(T-1):t}$ in the past $T$ time steps, our goal is to predict the future $\widehat{T}$ time steps as: $[X_{t-(T-1)}, ..., X_{t}]$ $\rightarrow$  $[X_{t+1}, ..., X_{t+\widehat{T}}]$, where $X_i$ $\in$ $\mathbb{R}^{N \times C}$ for the $i$-th time step, $N$ is the number of spatial nodes, and $C$ is the number of the information channel. Here C=1 in our datasets.

\begin{figure*}[t]
	\centering
        \includegraphics[width=1.0\textwidth]{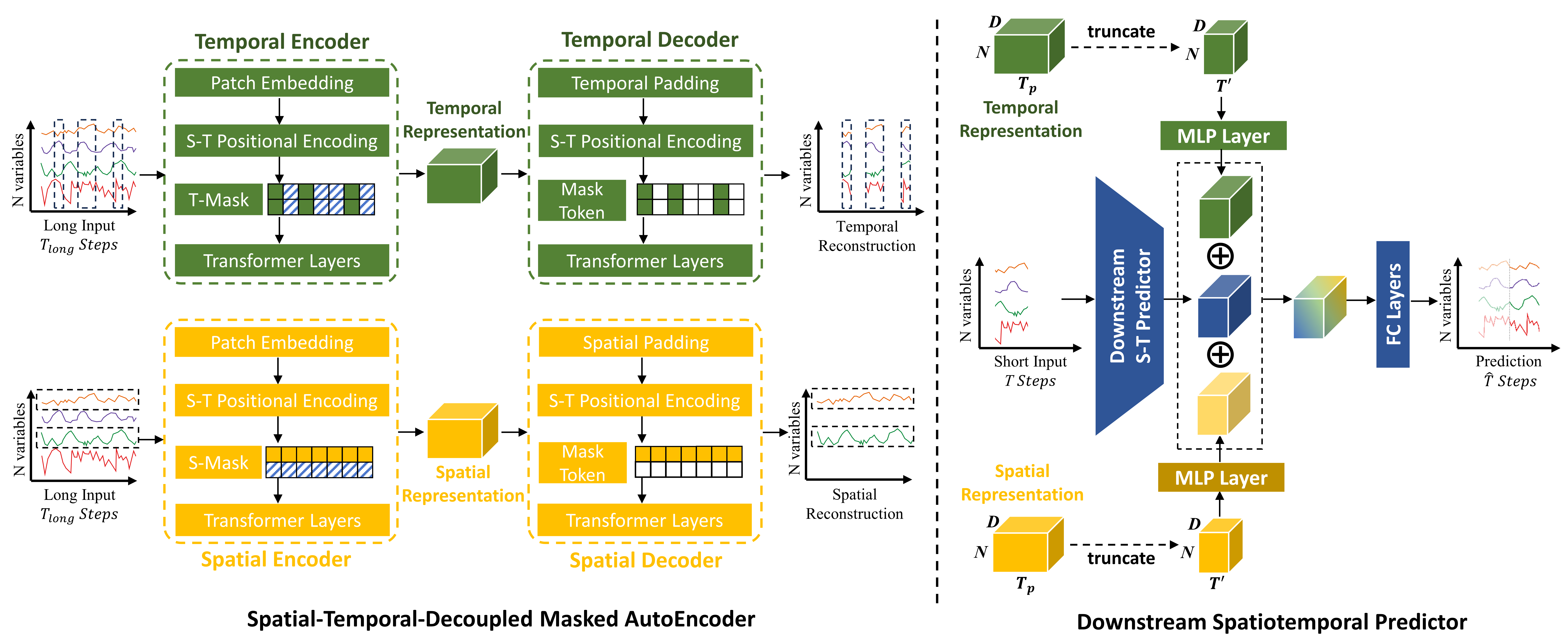}
	\caption{Spatial-Temporal-Decoupled Masked Pre-training Framework (\textbf{STD-MAE})}
	\label{fig:architecture}
\end{figure*}
\section{Methodology}
This section delves into the technical specifics of our proposed spatial-temporal-decoupled masked pre-training framework (\textbf{STD-MAE}) as delineated in Figure~\ref{fig:architecture}. 

\subsection{Spatial-Temporal Masked Pre-training}
\noindent\textbf{Spatial-Temporal-Decoupled Masking.} In the standard spatiotemporal forecasting task, the input length $T$ for $X_{t-(T-1):t}$ is usually equal to 12 (each step corresponds to 5-minute interval) ~\cite{yu2018spatio,wu2019graph,song2020spatial}, thus end-to-end models often fall into mirages outlined in Figure~\ref{fig1-sub3}. So we intend to introduce a masked pre-training phase involving long-range input. Since spatiotemporal data has an additional temporal dimension compared to image data and an extra spatial dimension over language data, a straightforward idea is to apply the original masked pre-training~\cite{kenton2019bert,he2022masked} directly by mixing the temporal and spatial dimension as one. However, this is unfeasible due to the square-level time and space complexity. Therefore, we propose a novel method called \textbf{\textit{spatial-temporal-decoupled masking}} during masked pre-training. This approach separately executes mask-reconstruction tasks along temporal and spatial dimension. Such decoupled masking mechanism allows the model to learn representation that can capture clearer heterogeneity. With this wider and clearer view, downstream predictors can see through spatiotemporal mirages. Consequently, our method presents an efficient and effective solution for pre-training on spatiotemporal data, enhancing model robustness against the challenges posed by complex spatiotemporal heterogeneity and mirage. 

Specifically, given an input spatiotemporal time series $\mathcal{X} \in \mathbb{R}^{T\times N \times C}$, we propose the following masking strategies: 
(1) \textbf{Spatial Masking (S-Mask)} randomly masks the time series of $N \times r$ sensors, where $r$ is the masking ratio between 0 and 1. This results in a spatially masked input $\widetilde{\mathcal{X}}^{(S)} \in \mathbb{R}^{T\times N(1-r) \times C}$. (2) \textbf{Temporal Masking (T-Mask)} randomly masks the time series of $T \times r$ time steps. This yields a temporally masked input $\widetilde{\mathcal{X}}^{(T)} \in \mathbb{R}^{T(1-r)\times N \times C}$. Both masking strategies can be viewed as random sampling from a Bernoulli distribution $\mathcal{B}(1-r)$ with expectation $1-r$ in the corresponding dimensions:
\begin{equation}
\begin{aligned}
\widetilde{\mathcal X}^{(S)} &= \sum_{n=1}^{N} \mathcal{B}_S(1-r) \cdot \mathcal X[:,n,:]\\
\widetilde{\mathcal X}^{(T)} &= \sum_{t=1}^{T} \mathcal{B}_T(1-r) \cdot \mathcal X[t,:,:]\\
\end{aligned}
\end{equation}
Intuitively, S-Mask forces the model to reconstruct the data of masked sensors solely from the other visible sensors, thus capturing long-range spatial heterogeneity. Similarly, temporal heterogeneity can be learned by utilizing the intrinsic visible series to reconstruct the entire time series with T-Mask.

\noindent\textbf{Spatial-Temporal-Decoupled Masked AutoEncoder.} Building upon the spatial-temporal-decoupled masking technique, we further propose the spatial-temporal-decoupled masked autoencoder. It consists of a temporal autoencoder (T-MAE) and a spatial autoencoder (S-MAE), both having a similar architecture. S-MAE applies self-attention along spatial dimension, while T-MAE performs self-attention along temporal dimension. Specifically, we consider long input with length $T_{long}$, typically spanning several days. However, directly utilizing such long sequences leads to computational and memory challenges. To address this, we apply a patch embedding technique~\cite{nie2022time}. The long input is divided into non-overlapping patches of length $T_p=T_{long}/L$ using a patch window $L$. This yields a patched input $\mathcal{X}_p\in\mathbb{R}^{T_p\times N \times LC}$. We then project $\mathcal{X}_p$ through a fully connected layer to obtain the patch embedding $E_{p}\in\mathbb{R}^{T_p\times N \times D}$, where $D$ is the embedding dimension. Moreover, to simultaneously encode spatial and temporal positional information, we implement a two-dimensional positional encoding~\cite{wang2021translating}. Given the patch embedding $E_{p}$, the spatiotemporal positional encoding $E_{pos}\in \mathbb{R}^{T_p\times N \times D}$ can be calculated as follows:
\begin{equation} \label{eq:pos}
	\begin{cases}
	{E_{pos}[t,n,2i]} = \sin(t/10000^{4i/D})\\
	{E_{pos}[t,n,2i\!+\!1]} = \cos(t/10000^{4i/D})\\
	{E_{pos}[t,n,2j\!+\!D/2]} = \sin(n/10000^{4j/D})\\
	{E_{pos}[t,n,2j\!+\!1\!+\!D/2]} = \cos(n/10000^{4j/D})
	\end{cases}
\end{equation}
We choose sinusoidal positional encoding instead of learned positional encoding because it can handle inputs of arbitrary length. The patch embedding $E_p$ and positional encoding $E_{pos}$ are summed to obtain the final input embedding $E\in \mathbb{R}^{T_p\times N \times D}$.

This input embedding $E$ is subsequently masked by S-Mask and T-Mask strategies to obtain the visible spatial patch embedding $\widetilde{E}^{(S)}$ and the visible temporal patch embedding $\widetilde{E}^{(T)}$. S-MAE and T-MAE generate spatial representations $H^{(S)}\in \mathbb{R}^{T_p\times N(1-r) \times D}$ and temporal representations $H^{(T)}\in \mathbb{R}^{T_p(1-r)\times N \times D}$ through a series of transformer layers, respectively. We denote $N_M=N\times r$ as the number of masked sensors and $T_M=T_p\times r$ as count of masked patches. By focusing only on visible parts, such design could reduce time and memory complexity.

A lightweight decoder is then applied to S-MAE and T-MAE to reconstruct the masked input. The spatial and temporal decoders each consists of a padding layer, a standard transformer layer, and a regression layer. Such asymmetrical design could dramatically reduce the pre-training time~\cite{he2022masked}. In the padding layer, we use a shared learnable mask token $V \in\mathbb{R}^{D}$ to indicate missing patches. Given spatial representation ${H}^{(S)}$ and temporal representation ${H}^{(T)}$, spatial padding expands $V$ to spatial mask tokens ${V}^{(S)} \in\mathbb{R}^{T_p \times N_M \times D}$ while temporal padding expands $V$ to temporal mask tokens ${V}^{(T)}\in\mathbb{R}^{T_M \times N \times D}$. The same spatiotemporal positional encoding as the encoders is added to ${V}^{(S)}$ and ${V}^{(T)}$. Then we perform concatenation operations respectively as $[{H}^{(S)}; {V}^{(S)}]$ and $[{H}^{(T)}; {V}^{(T)}]$ to get the full set of patches $\overline{H}^{(S)},\overline{H}^{(T)}\in\mathbb{R}^{T_p\times N \times L}$. Subsequently, they are passed to the transformer layer. Finally, a regression layer is used to reconstruct the time series at the patch level. Formally, the reconstruction $\widehat{Q}^{(S)}\in\mathbb{R}^{T_p\times N_M \times L}$ and $\widehat{Q}^{(T)}\in\mathbb{R}^{T_M \times N \times L}$ for the spatially and temporally masked inputs can be derived by:
\begin{equation}
	\begin{aligned}
\widehat{Q}^{(S)} =  M^{(S)} \odot FC(Attention^{(S)}(\overline{H}^{(S)}))\\
\widehat{Q}^{(T)} = M^{(T)} \odot FC(Attention^{(T)}(\overline{H}^{(T)}))
	\end{aligned}
\end{equation}
where ${M}^{(S)}\in\mathbb{R}^{N_M}$ is the spatial masked index and ${M}^{(T)}\in\mathbb{R}^{T_M}$ is the temporal masked index. 

Following other masked pre-training architectures~\cite{he2022masked,tong2022videomae}, we compute the loss on the masked part only. Our loss function is computed by calculating the mean absolute error (MAE) between the ground truth and the reconstruction result. Given reconstruction $\widehat{Q}^{(S)}$ and $\widehat{Q}^{(T)}$, the corresponding ground truth are $Q^{(S)}\in\mathbb{R}^{T_p\times N_M \times L}$ and $Q^{(T)}\in\mathbb{R}^{T_M\times N \times L}$. The two loss functions in spatial and temporal can be calculated as:
\begin{equation} \label{eq:mae-loss}
	\begin{aligned}
       	\mathcal{L}_S &= \frac{1}         {T_pN_{M}L}\mathop{\sum}_t^{T_{p}}\mathop{\sum}_n^{N_{M}}\mathop{\sum}_l^{L} \left|\widehat{Q}^{(S)}{[t,n,l]} - Q^{(S)}{[t,n,l]}\right|\\
    		\mathcal{L}_T &= \frac{1}{T_{M}NL}\mathop{\sum}_t^{T_{M}} \mathop{\sum}_n^{N}\mathop{\sum}_l^{L} \left|\widehat{Q}^{(T)}{[t,n,l]} - Q^{(T)}{[t,n,l]}\right| 
	\end{aligned}
\end{equation}
In summary, through the above spatial-temporal-decoupled masked pre-training, STD-MAE can capture clear and complete spatial and temporal heterogeneity.

\subsection{Downstream Spatiotemporal Forecasting}
STD-MAE can be seamlessly integrated into existing predictor structures. This operation is done by adding the spatial and temporal representations generated by STD-MAE to the hidden representation of the predictor. Concretely, we first feed long-range input with $T_{long}$ time steps into pre-trained spatial and temporal encoders to generate the corresponding spatial representation ${H}^{(S)}$ and temporal representation ${H}^{(T)}$. Then, we apply a downstream spatiotemporal predictor $\mathbb{F_\theta}$ with parameter $\theta$ to obtain the hidden representation $H^{(F)}$ $\in\mathbb{R}^{N \times D'}$ of the widely-used short input $X_{t-(T-1):t}$ through the following:
\begin{equation}
H^{(F)} = \mathbb{F_\theta}[X_{t-(T-1):t}]
\end{equation}
where $D'$ is the hidden representation dimension of the predictor. To align with $H^{(F)}$, we truncate the representation ${H}^{(S)}$ and ${H}^{(T)}$ of the last $T'$ patches, and reshape these two representations to ${H'}^{(S)}\in\mathbb{R}^{ N\times T'D}$ and ${H'}^{(T)}\in\mathbb{R}^ {N\times T'D}$. Next, we project these two representations into $D'$ dimension through a two-layer MLP. Finally, the augmented representation ${H^{(Aug)}}\in\mathbb{R}^{N \times D'}$ could be derived by adding these representations together:
\begin{equation}
{H^{(Aug)}} = MLP({H'}^{(S)})+MLP({H'}^{(T)})+H^{(F)}
\end{equation}
By far, ${H^{(Aug)}}$ includes representations generated by the predictor itself as well as the long-range spatial and temporal representations from STD-MAE, which can largely enhance the performance of the downstream spatiotemporal predictor. 

Specifically, in our work, we choose GWNet~\cite{wu2019graph} as our predictor due to its superior performance. We obtain the final representation by aggregating the hidden states from the skip connections across the multiple spatiotemporal layers of GWNet, along with the corresponding spatial and temporal representations generated by the STD-MAE. The augmented representation is then fed into GWNet's regression layers for prediction. Furthermore, we also test other classical spatiotemporal predictors with a variety of structures, i.e., DCRNN~\cite{li2018diffusion}, MTGNN~\cite{wu2020connecting}, STID~\cite{shao2022spatial} and STAEformer~\cite{liu2023spatio}. These experiments demonstrate the generality of STD-MAE. Details could be found in our ablation study. 

\section{Experiment}\label{sec:experiment-setup}
\subsection{Experimental Setup}
\noindent\textbf{Datasets.} To thoroughly evaluate the proposed STD-MAE model, we conduct extensive experiments on six real-world spatiotemporal benchmark datasets as listed in Table~\ref{tab:dataset}: PEMS03, PEMS04, PEMS07, PEMS08~\cite{song2020spatial}, METR-LA, and PEMS-BAY~\cite{li2018diffusion}. The raw data has a fine-grained time resolution of 5 minutes between consecutive time steps. For data preprocessing, we perform Z-score normalization on the raw inputs. 
\begin{table}[ht]
    \centering
    \small
    \begin{tabular}{cccc}
    \toprule
    Datasets & \#Sensors & \#Time Steps& Time Interval \\ 
    \midrule
    PEMS03    & 358  &5min     &26208  \\
    PEMS04     &307  &5min   & 16992 \\
    PEMS07    &883  &  5min&  28224\\ 
    PEMS08    &170  &  5min&  17856\\ 
    METR-LA    &207  &5min  & 34272 \\ 
    PEMS-BAY   &325  &5min  &  52116\\ 

    \bottomrule
    \end{tabular}
  \caption{Summary of Six Spatiotemporal Benchmarks}
  \label{tab:dataset}
\end{table}
\begin{table*}[h]
	\renewcommand{\arraystretch}{1.1}
	\centering
	\resizebox{1.0\linewidth}{!}{
	\begin{tabular}{l|c c c|c c c |c c c|c c c}
		\toprule[1.2pt]
		\multirow{2}{*}{Model} & \multicolumn{3}{c|}{PEMS03} & \multicolumn{3}{c|}{PEMS04} & \multicolumn{3}{c|}{PEMS07} & \multicolumn{3}{c}{PEMS08} \\
		{}& MAE & RMSE & MAPE & MAE & RMSE & MAPE & MAE & RMSE & MAPE & MAE & RMSE & MAPE \\
		\midrule
ARIMA~\cite{fang2021spatial}                & 35.31                & 47.59          & 33.78                      & 33.73                & 48.80             & 24.18                    & 38.17                &  59.27               & 19.46                & 31.09                &  44.32              & 22.73                 \\
VAR~\cite{song2020spatial}                  & 23.65                & 38.26          & 24.51                      & 23.75                & 36.66            & 18.09                    & 75.63                &  115.24              & 32.22                & 23.46                &  36.33              & 15.42                 \\
SVR~\cite{song2020spatial}                  & 21.97                & 35.29          & 21.51                      & 28.70                 & 44.56            & 19.20                     & 32.49                &  50.22               & 14.26                & 23.25                &  36.16              & 14.64                 \\
LSTM~\cite{song2020spatial}                 & 21.33                & 35.11          & 23.33                      & 27.14                & 41.59            & 18.20                     & 29.98                &  45.84               & 13.20                 & 22.20                 &  34.06              & 14.20                  \\
TCN~\cite{lan2022dstagnn}                   & 19.31                & 33.24          & 19.86                      & 31.11                & 37.25            & 15.48                    & 32.68                &  42.23               & 14.22                & 22.69                &  35.79              & 14.04                 \\
Transformer~\cite{vaswani2017attention}     & 17.50                 & 30.24          & 16.80                       & 23.83                & 37.19            & 15.57                    & 26.80                 &  42.95               & 12.11                & 18.52                &  28.68              & 13.66                 \\
DCRNN~\cite{li2018diffusion}                & 18.18                & 30.31          & 18.91                      & 24.70                 & 38.12            & 17.12                    & 25.30                 &  38.58               & 11.66                & 17.86                &  27.83              & 11.45                 \\
STGCN~\cite{yu2018spatio}                   & 17.49                & 30.12          & 17.15                      & 22.70                 & 35.55            & 14.59                    & 25.38                &  38.78               & 11.08                & 18.02                &  27.83              & 11.40                  \\
ASTGCN~\cite{guo2019attention}           & 17.69                & 29.66          & 19.40                       & 22.93                & 35.22            & 16.56                    & 28.05                &  42.57               & 13.92                & 18.61                &  28.16              & 13.08                 \\
GWNet~\cite{wu2019graph}                    & 19.85                & 32.94          & 19.31                      & 25.45                & 39.70             & 17.29                    & 26.85                &  42.78               & 12.12                & 19.13                &  31.05              & 12.68                 \\
STSGCN~\cite{song2020spatial}               & 17.48                & 29.21          & 16.78                      & 21.19                & 33.65            & 13.90                     & 24.26                &  39.03               & 10.21                & 17.13                &  26.80               & 10.96                 \\
STFGNN~\cite{li2021spatial}                 & 16.77                & 28.34          & 16.30                       & 19.83                & 31.88            & 13.02                    & 22.07                &  35.80                & 9.21                 & 16.64                &  26.22              & 10.60                  \\
STGODE~\cite{fang2021spatial}               & 16.50                 & 27.84          & 16.69                      & 20.84                & 32.82            & 13.77                    & 22.99                &  37.54               & 10.14                & 16.81                &  25.97              & 10.62                 \\
DSTAGNN~\cite{lan2022dstagnn}               & 15.57                & 27.21          & 14.68                      & 19.30                 & 31.46            & 12.70                     & 21.42                &  34.51               & 9.01                 & 15.67                &  24.77              & 9.94                  \\
ST-WA~\cite{cirstea2022towards}             & 15.17                & 26.63          & 15.83                      & 19.06                & 31.02            & 12.52                    & 20.74                &  34.05               & 8.77                 & 15.41                &  24.62              & 9.94                  \\
ASTGNN~\cite{guo2021learning}               & 15.07                & 26.88          & 15.80                       & 19.26                & 31.16            & 12.65                    & 22.23                &  35.95               & 9.25                 & 15.98                &  25.67              & 9.97                  \\
EnhanceNet~\cite{cirstea2021enhancenet}     & 16.05                & 28.33          & 15.83                      & 20.44                & 32.37            & 13.58                    & 21.87                &  35.57               & 9.13                 & 16.33                &  25.46              & 10.39                 \\
AGCRN~\cite{bai2020adaptive}                & 16.06                & 28.49          & 15.85                      & 19.83                & 32.26            & 12.97                    & 21.29                &  35.12               & 8.97                 & 15.95                &  25.22              & 10.09                 \\
Z-GCNETs~\cite{chen2021z}                   & 16.64                & 28.15          & 16.39                      & 19.50                 & 31.61            & 12.78                    & 21.77                &  35.17               & 9.25                 & 15.76                &  25.11              & 10.01                 \\
STNorm ~\cite{deng2021st}                  & 15.32                & 25.93          & 14.37                      & 19.21                & 32.30         & 13.05                    & 20.59                &  34.86               & 8.61                 & 15.39                &  24.80              & 9.91                  \\
STEP~\cite{shao2022pre}                     & 14.22                & 24.55          & 14.42                      & 18.20                 & 29.71            & 12.48                    & 19.32                &  32.19               & 8.12                 & 14.00                &  23.41              & 9.50                  \\
PDFormer~\cite{jiang2023pdformer}           & 14.94                    & 25.39              & 15.82                          & 18.32                & 29.97            & 12.10               & 19.83                &  32.87               & 8.53                 & 13.58                &  23.51              & 9.05                  \\
STAEformer~\cite{liu2023spatio}           & 15.35                    & 27.55             & 15.18                         & 18.22                & 30.18            & 11.98               & 19.14                &  32.60               & 8.01                & 13.46                &  23.25             & 8.88                  \\\hline
\textbf{STD-MAE (Ours)}                                    & \bf{13.80}           & \bf{24.43}     & \bf{13.96}                 & \bf{17.80}           & \bf{29.25}       & \bf{11.97}                    & \bf{18.65}           &  \bf{31.44}          & \bf{7.84} &          
\bf{13.44}           &  \bf{22.47}         & \bf{8.76}             \\
     \bottomrule[1.2pt]
     \end{tabular}
	}
	\caption{Performance Comparison with Baseline Models on PEMS03,04,07,08 Benchmarks}
	\label{tab:overall}
\end{table*}

\noindent\textbf{Baselines.} We compare STD-MAE with the following baseline methods. ARIMA~\cite{fang2021spatial}, VAR~\cite{song2020spatial}, SVR~\cite{song2020spatial}, LSTM~\cite{song2020spatial}, TCN~\cite{lan2022dstagnn}, and Transformer~\cite{vaswani2017attention} are time series models. For spatiotemporal models, we select several typical methods including DCRNN~\cite{li2018diffusion}, STGCN~\cite{yu2018spatio}, ASTGCN~\cite{guo2019attention},  GWNet~\cite{wu2019graph}, STSGCN~\cite{song2020spatial}, STFGNN~\cite{li2021spatial}, STGODE~\cite{fang2021spatial}, DSTAGNN~\cite{lan2022dstagnn}, ST-WA~\cite{cirstea2022towards}, ASTGNN~\cite{guo2021learning}, EnhanceNet \cite{cirstea2021enhancenet}, AGCRN~\cite{bai2020adaptive}, Z-GCNETs~\cite{chen2021z}, STEP~\cite{shao2022pre}, PDFormer~\cite{jiang2023pdformer} and STAEformer~\cite{liu2023spatio}. 

\noindent\textbf{Settings.} Following previous work~\cite{song2020spatial,li2021spatial,fang2021spatial,jiang2023pdformer,guo2021learning}, we divide the PEMS03, PEMS04, PEMS07, and PEMS08 datasets into training, validation, and test sets according to a 6:2:2 ratio. For METR-LA and PEMS-BAY datasets, the training, validation, and test ratio is set to 7:1:2. During pre-training, the long input $T_{long}$ of the six datasets are set to 864, 864, 864, 2016, 864, and 864 time steps, respectively. For prediction, we set the length of both input $T$ and output $\widehat{T}$ to 12 steps. The embedding dimension $D$ is 96. The encoder has 4 transformer layers while the decoder has 1 transformer layer. The number of multi-attention heads in transformer layer is set to 4. We use a patch size $L$ of 12 to align with the forecasting input. $T'$ is equal to 1, which means we truncate and keep the last one patch of ${H}^{(S)}$ and ${H}^{(T)}$. The masking ratio $r$ is set to 0.25. Optimization is performed with Adam optimizer using an initial learning rate of 0.001 and mean absolute error (MAE) loss. For evaluation, we use MAE, root mean squared error (RMSE), and mean absolute percentage error (MAPE(\%)). Performance in Table~\ref{tab:overall} is assessed by averaging over all 12 prediction steps. Experiments are mainly conducted on a Linux server with four NVIDIA GeForce RTX 3090 GPUs. To make fair and consistent comparison, they are all performed on BasicTS~\cite{shao2023exploring} platform.
\begin{table}[h]
    \renewcommand\arraystretch{1.2}
    \footnotesize
	\setlength{\tabcolsep}{0.8mm}{
	\begin{tabular*}{8.4cm}{@{\extracolsep{\fill}}lcccc}
		\hline
		\multirow{1}{*}{} & GWNet&STEP&PDFormer&\textbf{STD-MAE} \\
		\hline

Horizon@3 MAE& 1.30&1.26&1.32&\textbf{1.23}\\
Horizon@3 RMSE&2.73&2.73&2.83&\textbf{2.62}\\
Horizon@3 MAPE&2.71&2.59&2.78&\textbf{2.56}\\
		\hline
Horizon@6 MAE& 1.63&1.55&1.64&\textbf{1.53}\\
Horizon@6 RMSE&3.73&3.58&3.79&\textbf{3.53}\\
Horizon@6 MAPE&3.73&3.43&3.71&\textbf{3.42}\\
		\hline
Horizon@12 MAE& 1.99&1.79&1.91&\textbf{1.77}\\
Horizon@12 RMSE&4.60&4.20&4.43&\textbf{4.20}\\
Horizon@12 MAPE&4.71&4.18&4.51&\textbf{4.17}\\
		\hline
	\end{tabular*}}
\captionsetup{justification=centering}
		\caption{Performance on PEMS-BAY Dataset}
		\label{tab:bay}
\end{table}
\begin{table}[h]
    \renewcommand\arraystretch{1.2}
    \footnotesize
	\setlength{\tabcolsep}{0.8mm}{
	\begin{tabular*}{8.4cm}{@{\extracolsep{\fill}}lcccc}
		\hline
		\multirow{1}{*}{} & GWNet&STEP&PDFormer&\textbf{STD-MAE} \\
		\hline		
Horizon@3 MAE& 2.69&\textbf{2.61}&2.83&2.62\\
Horizon@3 RMSE&5.15&\textbf{4.98}&5.45&5.02\\
Horizon@3 MAPE&6.99&\textbf{6.60}&7.77&6.70\\
		\hline
Horizon@6 MAE& 3.08&\textbf{2.96}&3.20&2.99\\
Horizon@6 RMSE&6.20&\textbf{5.97}&6.46&6.07\\
Horizon@6 MAPE&8.47&\textbf{7.96}&9.19&8.04\\
		\hline
Horizon@12 MAE& 3.51&\textbf{3.37}&3.62&3.40\\
Horizon@12 RMSE&7.28&\textbf{6.99}&7.47&7.07\\
Horizon@12 MAPE&9.96&9.61&10.91&\textbf{9.59}\\
		\hline
	\end{tabular*}}
		\caption{Performance on METR-LA Dataset}
		\label{tab:la}
\end{table}
\subsection{Overall Performance}\label{sec:experiment-overall}
The performance of models is listed in Table~\ref{tab:overall}, Table~\ref{tab:bay}, and Table~\ref{tab:la}. For a fair comparison, the reported results of the baseline models are taken from the original literature, which have been widely cited and validated in spatiotemporal forecasting. Across all datasets, our STD-MAE achieves superior performance over the baselines by a significant margin on all evaluation metrics. For other baselines, the spatiotemporal models clearly outperform the time series models due to their ability to capture spatiotemporal dependencies. In summary, the proposed STD-MAE framework significantly advances the state-of-the-art in spatiotemporal forecasting, demonstrating its ability to augment downstream predictors.

\subsection{Ablation Study}
\noindent\textbf{Masking Ablation.} We design four variants to validate the effectiveness of our spatial-temporal masking mechanism:
\begin{itemize}
  \item \textbf{S-MAE}: Only masking on the spatial dimension.
  \item \textbf{T-MAE}: Only masking on the temporal dimension.
  \item \textbf{STM-MAE}: Using spatial-temporal-mixed masking.
  \item \textbf{w/o Mask}: Without applying any masked pre-training.
\end{itemize}  
\begin{figure}[h]
  \centering
    \begin{subfigure}{0.95\linewidth}
    \includegraphics[width=\linewidth]{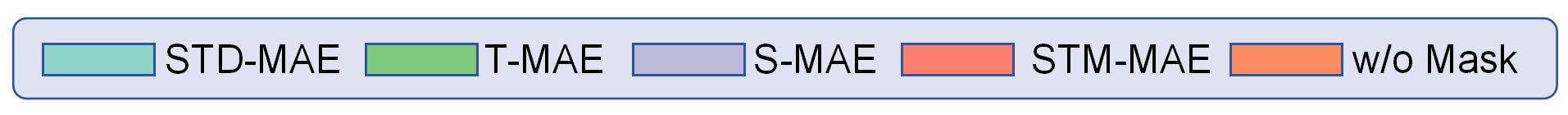}
  \end{subfigure}
     \begin{subfigure}{0.95\linewidth}
    \includegraphics[width=\linewidth]{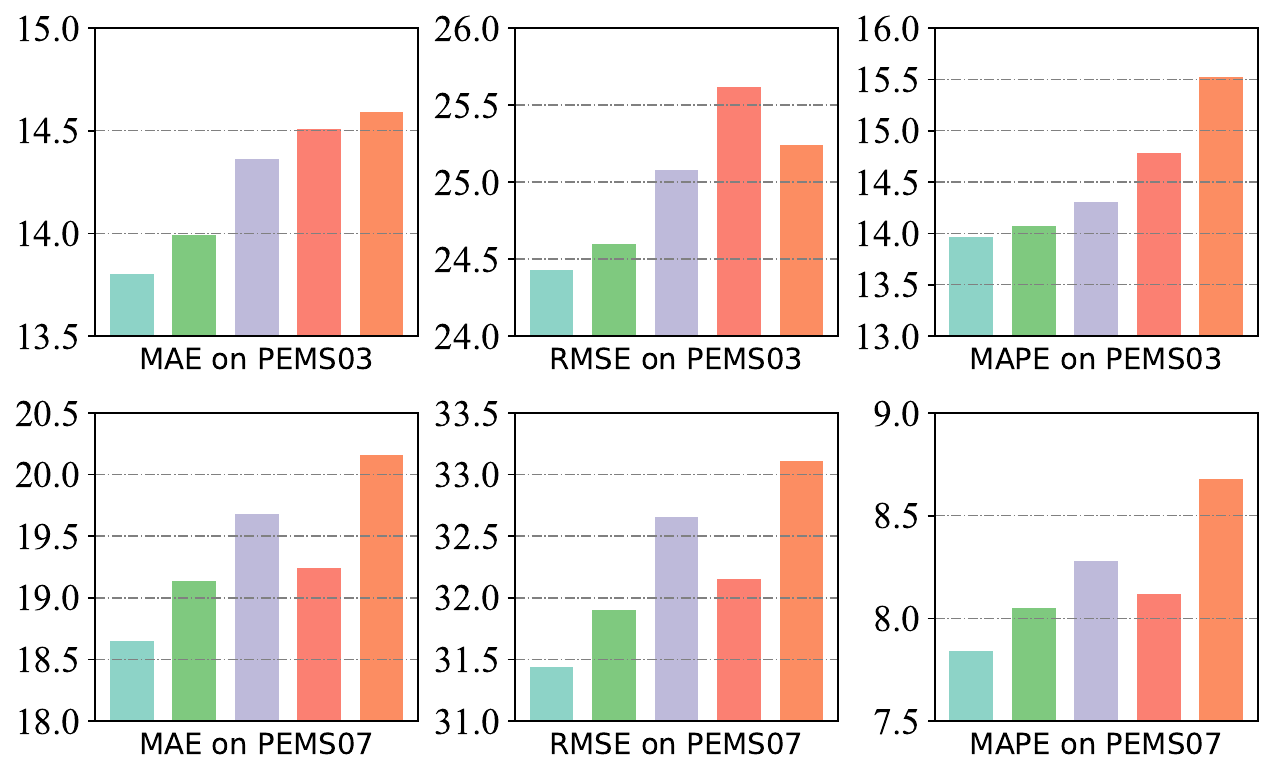}
  \end{subfigure}
  \caption{Masking Ablation on PEMS03 and PEMS07}
  \label{fig:ablation1}
\end{figure}
We report the experimental results on the PEMS03 and PEMS07 datasets. As illustrated in Figure~\ref{fig:ablation1}, STD-MAE with spatial-temporal-decoupled masking significantly outperforms the ablated versions. T-MAE and S-MAE still improve over the original model although they can only partly capture the heterogeneity. For STM-MAE, we mix the spatial and temporal dimensions before randomly masking operation. The task of mixed masking is trivial which would lead to learn representation with less rich semantics. Overall, the results highlight the value of our proposed spatial-temporal-decoupled masked pre-training design for spatiotemporal forecasting.

\noindent\textbf{Predictor Ablation.} To evaluate the generality of STD-MAE, we test five downstream predictors with different backbones including GCN+RNN, GCN+TCN, Linear and Transformer:
\begin{itemize}
  \item \textbf{STD-MAE-DCRNN}: Using DCRNN as the predictor.
  \item \textbf{STD-MAE-MTGNN}: Using MTGNN as the predictor.
  \item \textbf{STD-MAE-STID}: Using STID as the predictor.
  \item \textbf{STD-MAE-STAE}: Using STAEformer as the predictor.
  \item \textbf{STD-MAE}: Using GWNet as the predictor.
 \end{itemize}
The experiments are conducted on the PEMS04 and PEMS08 datasets. Table~\ref{tab:ablationstudy2} illustrates consistent and substantial performance gains across all five downstream spatiotemporal predictors when augmented with STD-MAE. This demonstrates the robustness of the representations generated by STD-MAE, which can benefit all kinds of downstream predictors regardless of architectures.

In conclusion, these ablation studies validate the effectiveness of STD-MAE's decoupled design and the generality for enhancing downstream baselines.
\begin{table}[t]
    \footnotesize
    \centering
	\setlength{\tabcolsep}{0.8mm}{
	\begin{tabular*}{8.4cm}{@{\extracolsep{\fill}}lcc}
		\hline
		\multirow{2}{*}{Model} & PEMS04 & PEMS08 \\
            \cline{2-3}
		\multicolumn{1}{l}{} & 
		\multicolumn{1}{c}{MAE/RMSE/MAPE} & 
		\multicolumn{1}{c}{MAE/RMSE/MAPE}  \\
		\hline
DCRNN&19.63/31.24/13.52&15.21/24.11/10.04\\
STD-MAE-DCRNN&\bf{18.65/30.09/13.07}&\bf{14.50/23.38/9.36}\\
\hline
MTGNN&19.17/31.70/13.37&15.18/24.14/10.20\\
STD-MAE-MTGNN&\bf{18.72/31.03/12.72}&\bf{14.84/23.58/9.58}\\
\hline
STID&18.35/29.86/12.50  &14.21/23.35/9.32
\\
STD-MAE-STID&\bf{17.93/29.43/12.11}&\bf{13.53/22.60/8.97}
\\
\hline
STAEformer&18.22/30.18/\textbf{11.98} &13.46/23.25/8.88
\\
STD-MAE-STAE&\textbf{17.92/29.37/}12.11&\bf{13.30/22.51/8.82
}
\\\hline
GWNet&18.74/30.32/13.10&14.55/23.53/9.31\\
\textbf{STD-MAE (Ours)} &\bf{17.80/29.25/12.97}&\bf{13.44/22.47/8.76}\\
		\hline
	\end{tabular*}}
		\caption{Predictor Ablation on PEMS04 and PEMS08}
		\label{tab:ablationstudy2}
\end{table}
\subsection{Hyper-parameter Study}
\begin{figure}[t]
  \centering
    \includegraphics[width=0.95\linewidth]{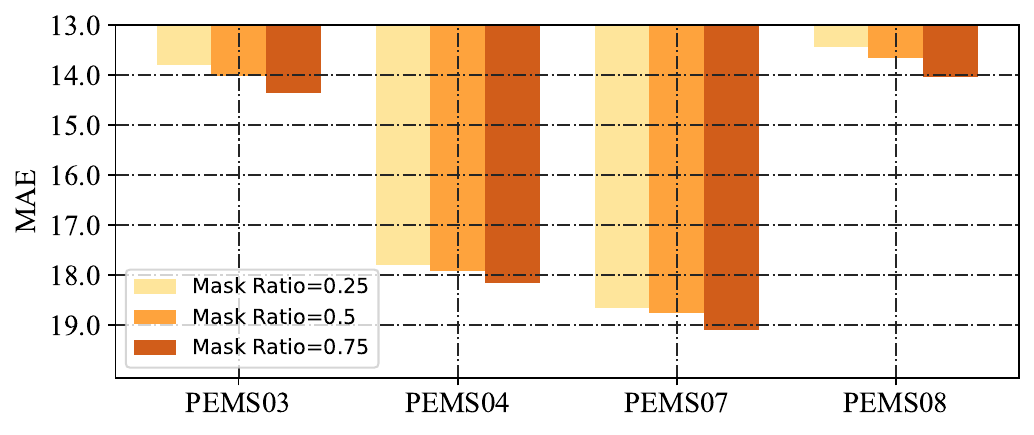}
  \caption{Hyper-parameter Study on Masking Ratio}
  \label{fig:hyper1}
\end{figure}

\noindent\textbf{Masking Ratio.} We first conduct hyperparameter study by varying the masking ratio $r$ in $\{0.25,0.5,0.75\}$, where the value is applied equally to S-MAE and T-MAE. As shown by Figure~\ref{fig:hyper1}, a masking ratio of 0.25 yields the lowest error across all datasets, indicating optimal value. Prior work on masked language modeling with BERT~\cite{kenton2019bert} utilizes a relatively low masking ratio of only 15\% during pre-training. In contrast, masked autoencoders for image reconstruction~\cite{he2022masked} and video modeling~\cite{tong2022videomae} have found much higher optimal ratio of 75\% and 90\%, respectively. However, we find a somewhat 
lower optimal ratio of 25\% for spatiotemporal time series modeling due to the fact that the long input time series required to provide extensive temporal context. This presents a challenge for reconstructing masked inputs, especially with spatial masking patterns that obstruct large contiguous blocks. Furthermore, our study demonstrates the importance of tuning masking specifically for spatiotemporal data. While an exact optimal is dataset-dependent, our results nonetheless show that relatively lower masking ratio is preferable for spatiotemporal time series.

\noindent\textbf{Pre-training Length.} We also study the effects of $T_{long}$, which is the input length used in pre-training. Here we test different pre-training lengths of one day, three days, and a week on all four datasets. The results are shown in Table~\ref{tab:hyper2}. In two out of the four datasets, a pre-training length of 3 days yields the best performance. Remarkably, compared to previous pre-training methodology, our approach demonstrates an enhanced capability to achieve superior performance with comparatively shorter pre-training lengths. These findings not only show the dynamic impact of pre-training length on the performance but also guide that the optimal pre-training length changes according to datasets.
\begin{table}[t] 
    \footnotesize
    \centering
	\begin{tabular}{lccc}
            \toprule
		\multirow{2}{*}{$T_{long}$} & 288 &864 & 2016 \\
		\cline{2-4}
		\multicolumn{1}{l}{} & 
		\multicolumn{1}{c}{MAE / RMSE} & 
		\multicolumn{1}{c}{MAE / RMSE} & 
		\multicolumn{1}{c}{MAE / RMSE} \\
		\midrule
PEMS03&13.82/24.97&\bf{13.80/24.43}&14.07/25.01\\
PEMS04&17.94/29.26&\bf{17.80/29.26}&17.84/29.27\\
PEMS07&19.01/31.85&18.65/31.44&\bf{18.31/31.07}\\
PEMS08&13.74/22.71&13.66/22.68&\bf{13.44/22.47}\\
		\bottomrule
	\end{tabular}
	\caption{Hyper-parameter Study on Pre-training Length $T_{long}$}
	\label{tab:hyper2}
\end{table}

\subsection{Efficiency Test}
Since STD-MAE introduce two decoupled autoencoders to encode spatial and temporal representation, efficiency might be a concern. However, due to the decoupled design of STD-MAE, it still outperforms in efficiency compare to other pretrained model especially for datasets with a large amount of sensors. Specifically, compared to non-decoupled pre-training methods, our decoupled time complexity is $O(N^2+T_{p}^2)$. We report the total training time for pre-training and forecasting per sample of four datasets as Table~\ref{tab:effc}.
\begin{table}[h]
    \centering
    \footnotesize
	\begin{tabular}{ccccc}
		\hline
		\multirow{1}{*}{} &\# Sensors&STEP&\textbf{STD-MAE}&\textit{Improvement} \\
		\hline
PEMS03&358&108ms&\textbf{50ms}&\textit{53.7\%}\\
PEMS04&307&73ms&\textbf{34ms}&\textit{53.9\%}\\
PEMS07&883&516ms&\textbf{142ms}&\textit{72.5\%}\\
PEMS08&170&62ms&\textbf{48ms}&\textit{22.6\%}\\
		\hline
	\end{tabular}
     \caption{Efficiency Test with Pre-training Models}
     \label{tab:effc}
\end{table}
\subsection{Case Study}
\begin{figure}[h]
  \centering
  \begin{subfigure}{0.45\textwidth}
    \includegraphics[width=\linewidth]{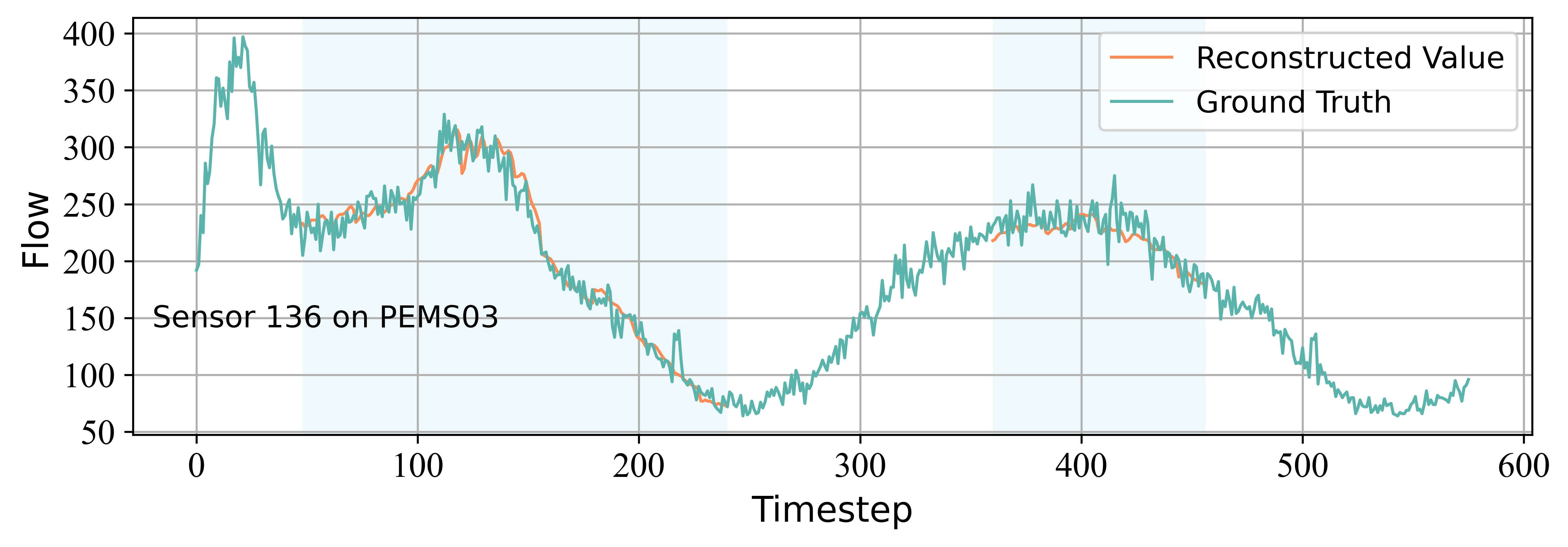}
        \caption{Reconstruction from T-MAE Pre-training}
    \label{case1-subfig1}
  \end{subfigure}
  \begin{subfigure}{0.45\textwidth}
    \includegraphics[width=\linewidth]{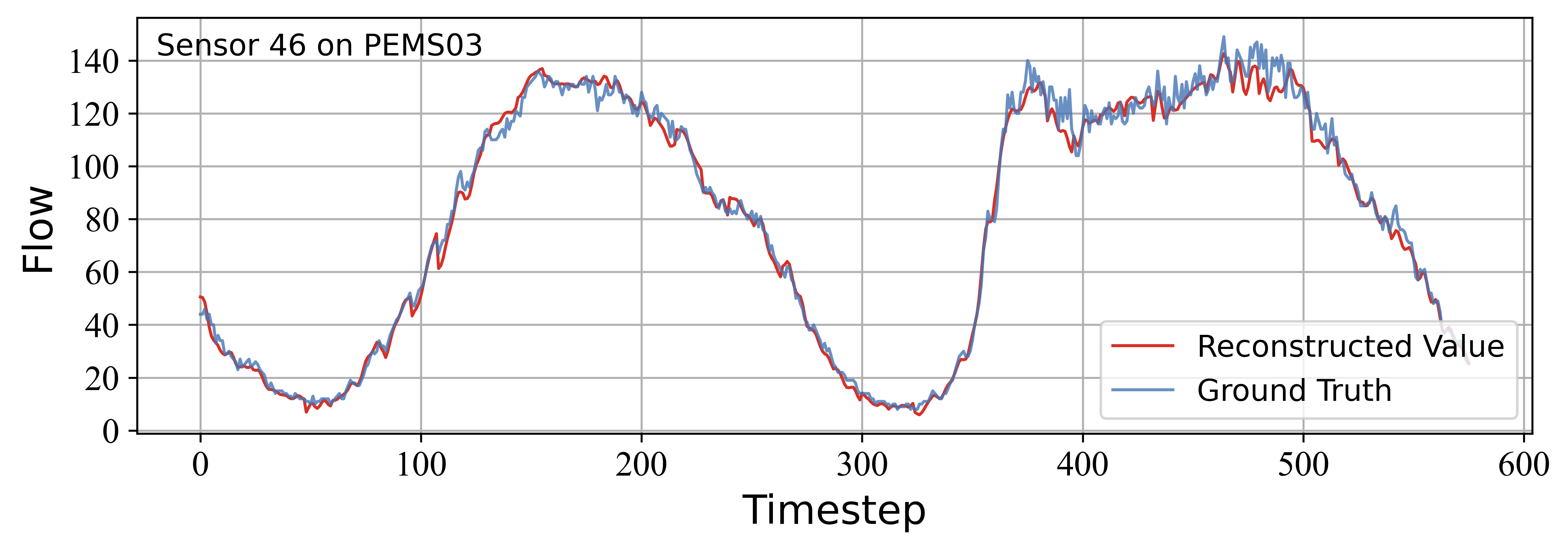}
        \caption{Reconstruction from S-MAE Pre-training}
    \label{case1-subfig2}
  \end{subfigure}
  \caption{Reconstruction Accuracy from Pre-training}
  \label{fig:case1}
\end{figure}

\noindent\textbf{Reconstruction Accuracy in Pre-training.} STD-MAE exhibits a robust capacity for reconstructing long time series by relying solely on the partially observed sensor recordings, as shown in Figures~\ref{case1-subfig1} and \ref{case1-subfig2}. The shaded area indicates the masked region. Temporal reconstruction closely matches the ground truth in terms of periodicity and trends. This suggests STD-MAE successfully acquires generalized knowledge of temporal patterns. Similarly, STD-MAE could restore entirely masked sensors based on contextual data from spatially related sensors, indicating it also gains meaningful spatial correlations. Overall, STD-MAE appears to learn rich spatiotemporal representations through pre-training.

\begin{figure}[h]
    \centering
    \begin{subfigure}[b]{0.48\linewidth}
        \includegraphics[width=\linewidth]{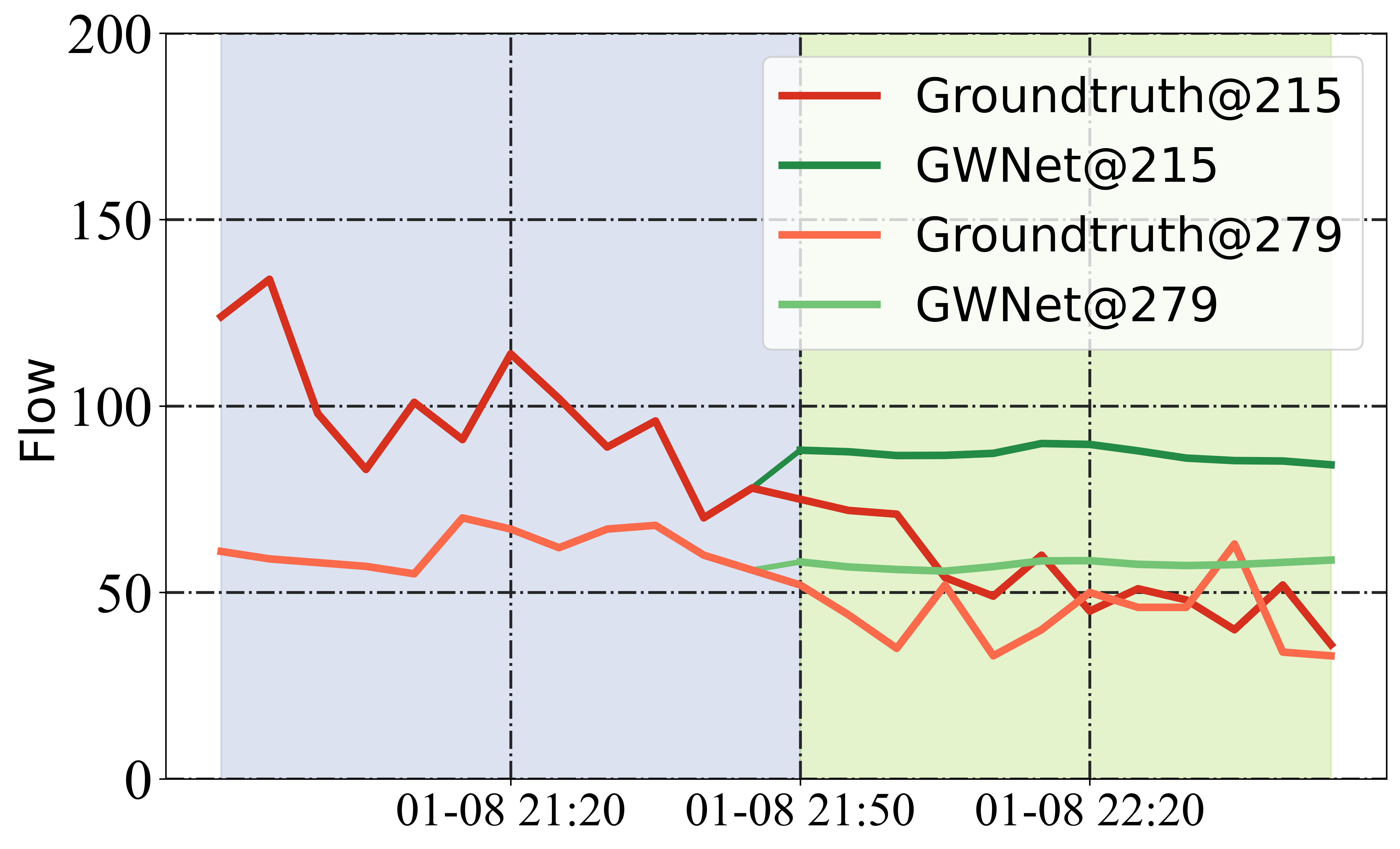}
        \caption{GWNet's Prediction}
        \label{fig:case2a}
    \end{subfigure}
    \begin{subfigure}[b]{0.48\linewidth}
        \includegraphics[width=\linewidth]{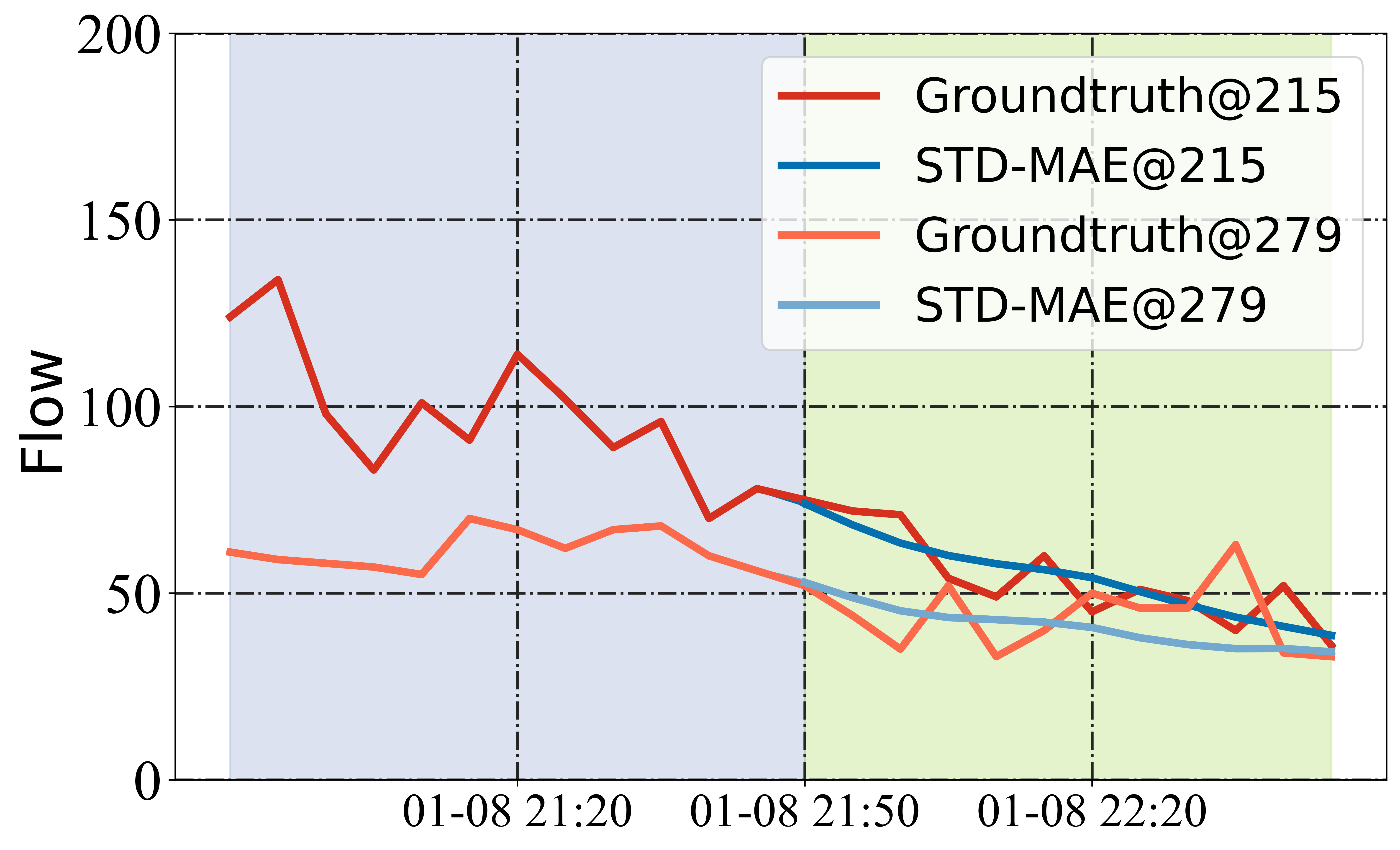}
        \caption{STD-MAE's Prediction}
        \label{fig:case2b}
    \end{subfigure}
    \begin{subfigure}[b]{0.48\linewidth}
        \includegraphics[width=\linewidth]{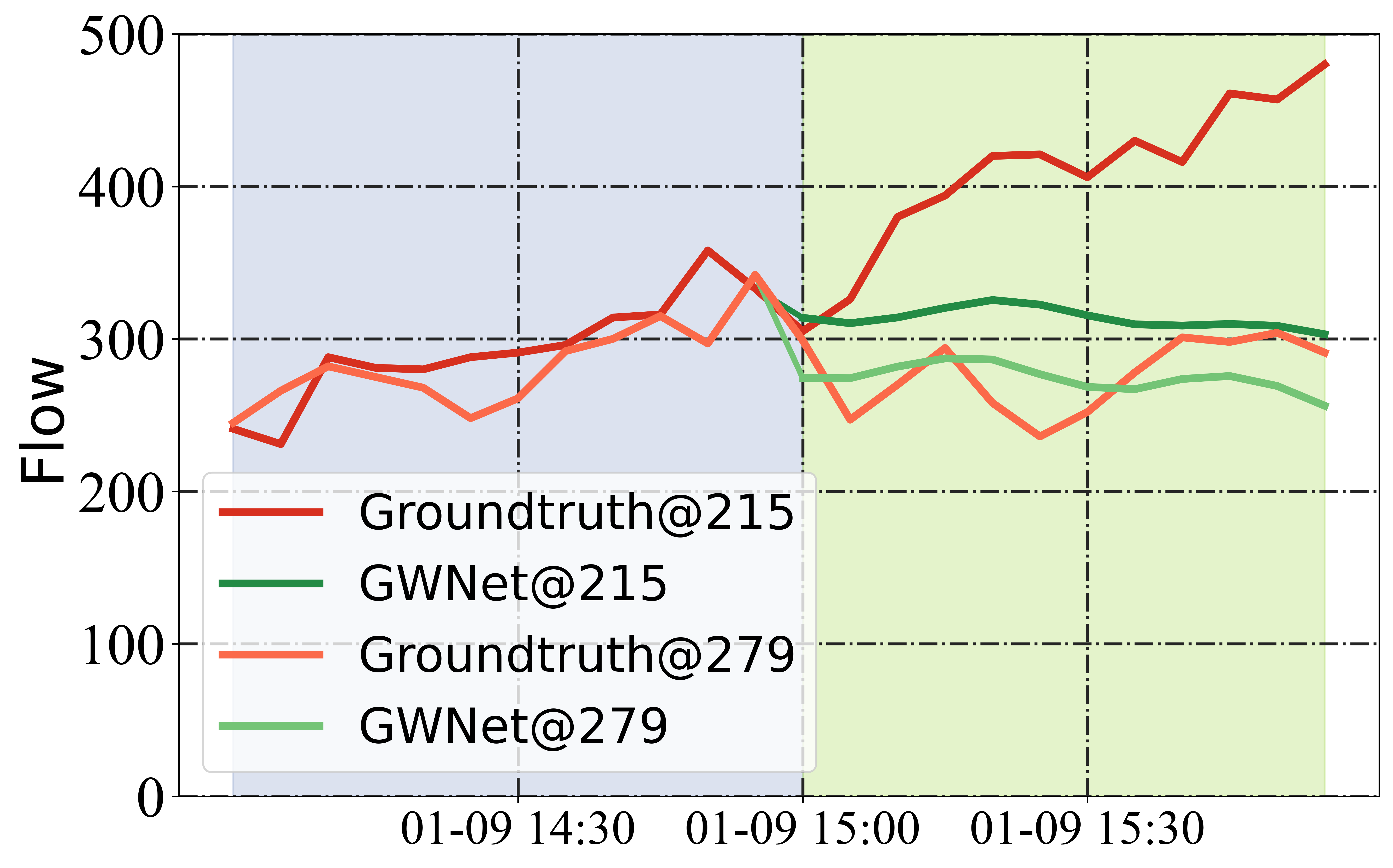}
        \caption{GWNet's Prediction}
        \label{fig:case2c}
    \end{subfigure}
    \begin{subfigure}[b]{0.48\linewidth}
        \includegraphics[width=\linewidth]{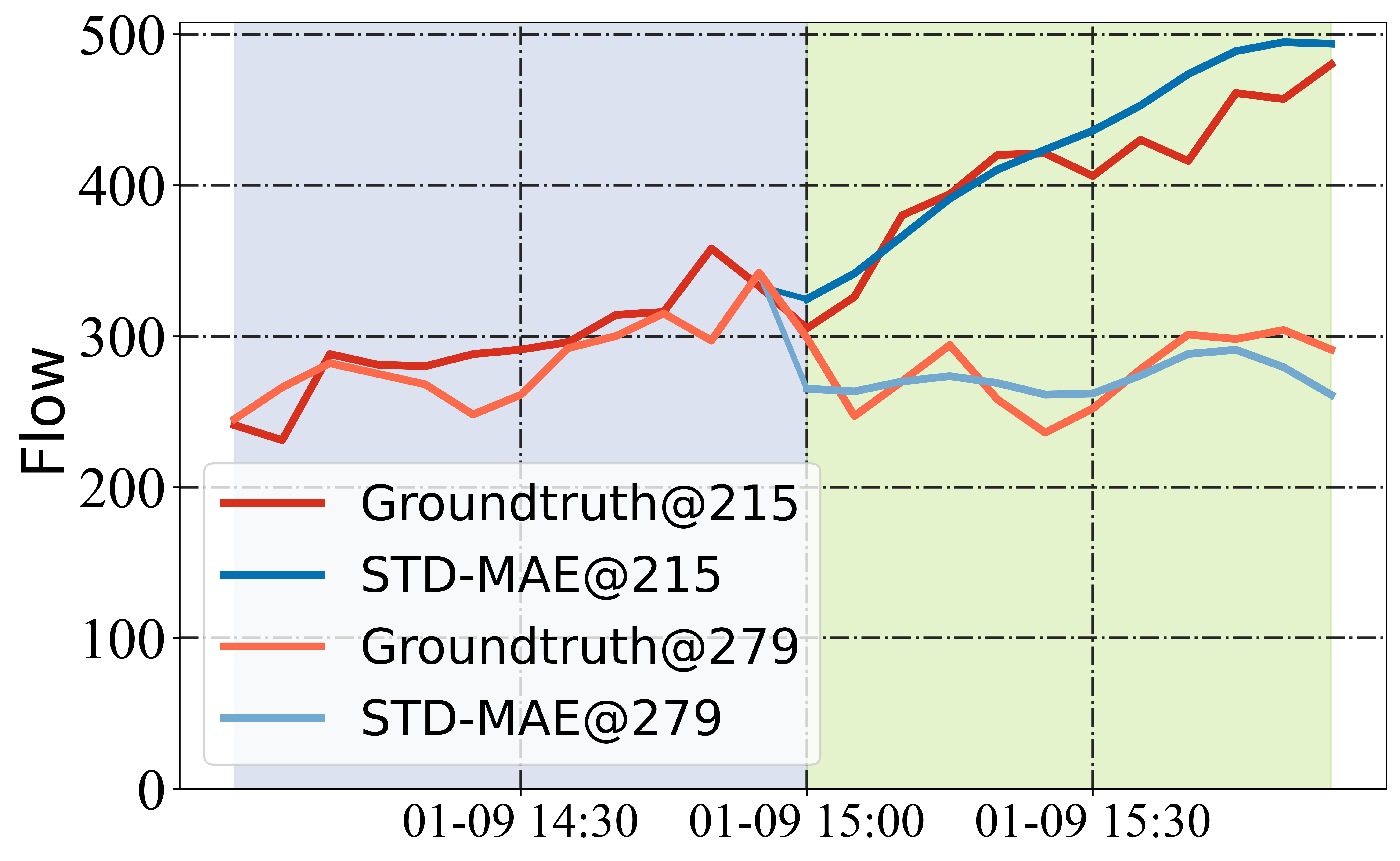}
        \caption{STD-MAE's Prediction}
        \label{fig:case2d}
    \end{subfigure}
    \caption{Prediction under Spatiotemporal Mirage}
    \label{fig:case2}
\end{figure}

\noindent\textbf{Robustness on Spatiotemporal Mirage.} In Figure~\ref{fig:case2}, we present a comparative analysis of the prediction results from GWNet and STD-MAE for two spatiotemporal mirages in Figure~\ref{fig1-sub3}. The input and prediction windows are denoted by purple and green backgrounds, respectively. A pivotal finding is that GWNet exhibits a limitation in distinguishing spatiotemporal mirages in Figures~\ref{fig:case2a} and~\ref{fig:case2c}. In contrast, STD-MAE performs a significant accuracy in these situations as shown in Figures~\ref{fig:case2b} and~\ref{fig:case2d}. The pre-trained component in STD-MAE remarkably enhances GWNet's ability to distinguish spatiotemporal mirages arising from heterogeneity.

\section{Conclusion}\label{sec:conclusion}
In this study, we propose STD-MAE, a novel spatial-temporal-decoupled masked pre-training framework for spatiotemporal forecasting. In the pre-training phase, a novel spatial-temporal-decoupled masking approach is utilized to effectively model the heterogeneity of spatiotemporal data. For forecasting phase, the hidden representations generated by STD-MAE are leveraged to boost the performance of downstream spatiotemporal predictors. Comprehensive experiments and in-depth analyses conducted on six benchmark datasets demonstrate the superiority of STD-MAE.
\bibliographystyle{named}
\section*{Acknowledgment}
This work was supported by JST CREST Grant Number JPMJCR21M2 including AIP challenge program and JSPS KAKENHI Grant Number JP24K02996.
\bibliography{ijcai24}
\end{document}